# Stable Stair-Climbing of a Quadruped Robot

Ali Zamani
Dept. of Mech. Eng.
K. N. Toosi Univ. of Technology
Tehran, Iran.
*ali.zamani1986 @gmail.com*

Mahdi Khorram
Dept. of Mech. Eng.
K. N. Toosi Univ. of Technology
Tehran, Iran.
*mahdi.khorram@gmail.com*

S. Ali A. Moosavian
Dept. of Mech. Eng.
K. N. Toosi Univ. of Technology
Tehran, Iran.
*moosavian@kntu.ac.ir*

*Abstract* — Synthesizing a stable gait that enables a quadruped robot to climb stairs is the focus of this paper. To this end, first a stable transition from initial to desired configuration is made based on the minimum number of steps and maximum use of the leg workspace to prepare the robot for the movement. Next, swing leg and body trajectories are planned for a successful stair-climbing gait. Afterwards, a stable spinning gait is proposed to change the orientation of the body. We simulate our gait planning algorithms on a model of quadruped robot. The results show that the robot is able to climb up stairs, rotate about its yaw axis, and climb down stairs while its stability is guaranteed.

*Index Terms*—Stability, wave gait, spinning gait, quadruped robot, transition, stair-climbing

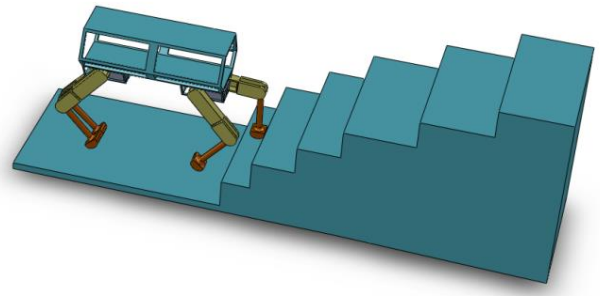

Fig. 1. Model of the quadruped robot.

## I. INTRODUCTION

Research in the field of legged robots has made great strides in the last decade and is one of the active fields of robotics. The main purpose of developing legged robots is that they possess high capabilities of traversing and exploring places which are inaccessible to wheeled or tracked vehicles. Despite significant progress in this field in recent years [1-4], the capabilities of legged robots developed hitherto are still far less than those of their biological counterparts.

One of the main issues of legged robots is to synthesize stable gaits. A legged robot is statically stable if the projection of its center of mass on the horizontal plane is within the support polygon formed by the feet on the ground. The minimum distance from the projection of the center of mass to the sides of the support polygon is referred to as stability margin. If the stability margin is positive, the robot is statically stable, otherwise unstable. Song and Waldron [5] defined gait as follow: "A gait is defined by the time and the location of the placing and lifting of each foot, coordinated with the motion of the body in its six degrees of freedom, in order to move the body from one place to another". Gaits can be classified as periodic gaits and aperiodic gaits. In periodic gaits, the different movement components occur at the same instants in a locomotion cycle. However, in aperiodic gaits, the different movement components are planned in a flexible manner as a function of the trajectory, the ground features and the machine's states. Periodic gaits are useful for moving on regular terrains as they provide high speed and optimal stability margin, but they require fixed starting foot positions with respect to the body to perform the movement. On the other hand, aperiodic gaits are suitable for moving on terrains with sharp irregularities, but their implementation is complicated. We use periodic gaits in this paper.

Stairs are specific types of terrains that legged robots sometime need to overcome. Based on the number of stairs that each leg advances in a gait cycle, there are generally two approaches for stair climbing. If each leg advances only one stair in each cycle, the gait support patterns will be skew symmetrical about the body longitudinal axis and thus gait stability is reduced [6]. Accordingly, the robot velocity is decreased. This method is appropriate when small leg workspace is available. On the other hand, if each leg advances two stairs in a gait cycle, the gait support patterns are symmetrical about the body longitudinal axis. As a result, the robot's stability and velocity are increased. In this paper we use the latter approach for stair climbing.

Stair-climbing gaits have been investigated by some researchers. Fu and Chen developed walking control for a humanoid robot to realize stable and robust stair climbing [7]. Sato et al. have proposed the virtual slope method for the walking trajectory planning on stairs for biped robots [8]. This method solves two problems about the Zero Moment Point (ZMP), namely the ZMP equation problem on stairs and the ZMP definition problem in a double-support phase. Song developed a walking chair for the disabled and studied some fundamental issues of a practical walking chair [6].

In previous studies, it is assumed that the robot's feet are in the desired positions in the body reference frame at the beginning of a periodic gait. However, here we propose a method for stable transitioning to the desired positions. This transition is based on the minimum number of steps and the maximum use of the leg workspace. The model we consider in this paper is a quadruped robot shown in Fig. 1. We synthesize stair-climbing and spinning gaits that enable the robot to climb stairs and rotate about its yaw axis while the stability of the robot is guaranteed. The paper is organized as follows. In section II, we describe necessary tools for creating the gaits. Section III presents the simulation results and section IV presents the conclusions.

## II. METHODS

*A. Stair-climbing gait*

Periodic gaits require fixed starting foot positions (desired configuration) in the body reference frame to perform the movement, and these positions depend on the trajectory that the periodic gait follows. We suppose that the robot is first in the initial configuration where each foot is positioned in the center of its workspace, which is assumed to be in a cube form, with respect to the body. We also consider a continuous wave gait, a periodic symmetric gait in which the placement of each foot starts from the back leg and advances toward the front leg like a wave along each side of the body. Thus, for a continuous wave gait in the positive x-direction, the swing-leg sequence is $1 - 4 - 2 - 3$ (see Fig. 2). To exploit the maximum workspace of each leg to achieve high speed, each foot with respect to the body at the beginning of the wave gait should be appropriately positioned in such a way that each leg advances the maximum leg stroke during the locomotion. We call the appropriate foothold positions as desired configuration as shown by solid circles in Fig. 2.

To initiate the wave gait, a stable transition from the initial configuration to the desired configuration is required. This transition is illustrated in Fig. 3. The solid circles depict the foothold positions during the support phase and the open circles depict the final positions of the swing legs. The transition is carried out in five steps. Legs *2*, *3*, *4*, *1*, *3* are respectively moved in steps *1*, *2*, *3*, *4*, *5* while the body is stationary. As seen, the final foothold positions at step 5 are the same as those (desired configuration) shown in Fig. 2 and the projection of the center of mass lies inside or on the border of the support polygon at each step. This indicates a stable transition from the initial to configuration for the wave gait.

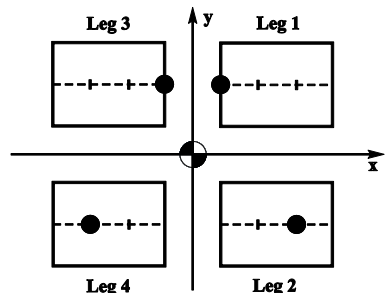

Fig 2. The desired foot positions with respect to the body at the beginning of wave gait.

We synthesize a stable gait that enables the robot to climb stairs. Since the dimensions of the stairs are assumed to be fixed, periodic gaits such as wave gaits can be employed. During stair-climbing, the robot body can be either leveled or tilted depending on the task. We consider the former one in this paper.

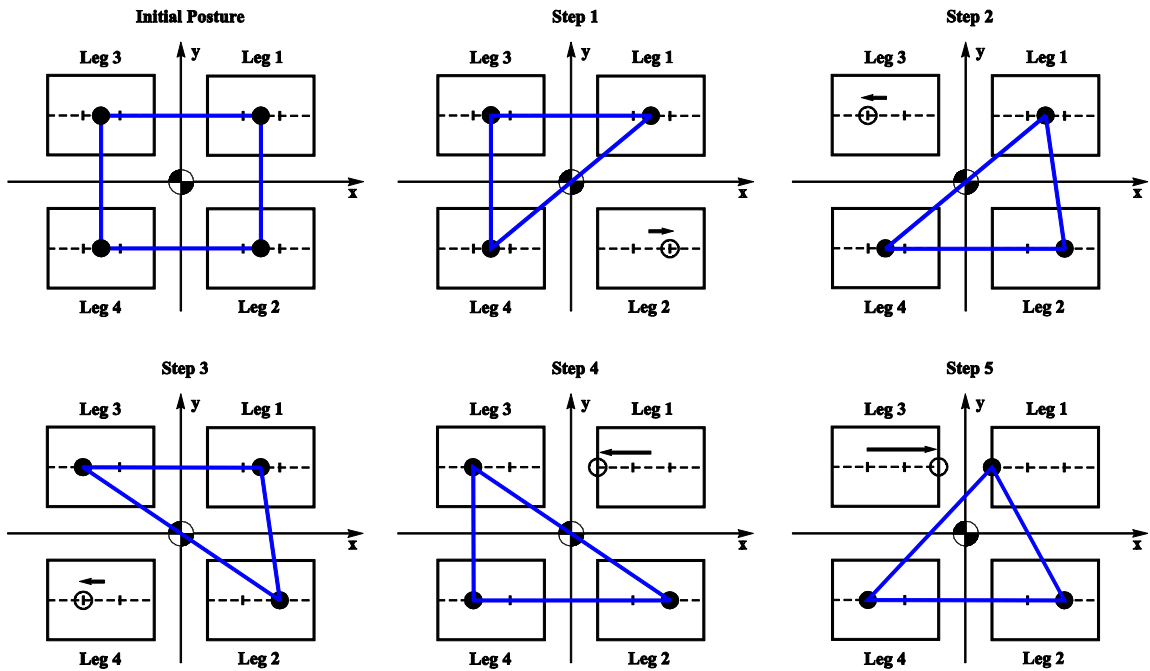

Fig. 3. Transition from the initial to desired configuration for the waive gait

Fig. 4 shows the robot ascending stairs. Since the distance between the footprints of any leg in successive cycle (footprint of leg *i*th in cycle *n* and footprint of leg *i*th in cycle *n+1*) is $\lambda=R/\beta$ [6], and each leg advances two stairs in one cycle, the minimum length of leg stroke, *R*, is *2Wβ* and *2Hβ* in horizontal and vertical directions, respectively, where *H* and *W* are the height and the length of each stair, and *β* is the duty factor.

We use the wave gait with swing leg sequence 1 – 4 – 2 – 3 for climbing stairs and each leg advances two stairs. For example, leg *4* is lifted from stair 1 and will be placed at stair 3 and, leg 3 is lifted from stair *2* and will be places at stair *4*, and so on. During the movement of each leg, the body translates a distance of $\left(\frac{1}{\beta}-1\right)R$ with constant speed in the *x* and *z* directions. Since we consider the case in which the body remains leveled relative to the ground during stair climbing, its stability condition is the same as walking on the level ground.

To avoid collision between the swing leg and stairs, the trajectory of the swing leg should be planned carefully, and the trajectory should be smooth during the swing phase. As such we consider a fifth-order polynomial for the trajectory of the swing leg in *x* and *z* directions. To avoid the collision, the *z* component of the foot trajectory at time $t_s$, corresponding to the *x* component of the foot trajectory being $d_s$, should be greater than the height of the stair as *Δh*, which is a small positive number (see Fig. 5). To minimize the effect of impact during the foot strike with stairs, initial (takeoff) and final (touchdown) velocities and accelerations of the swing leg trajectories in the *x* and *z* directions are set to zero. Thus, the swing leg trajectory in the *x* direction is given by

$$x = \frac{6x_f}{T_{sw}^5}t^5 - \frac{15x_f}{T_{sw}^4}t^4 + \frac{10x_f}{T_{sw}^3}t^3 \qquad (1)$$

where $x_f$ and $T_{sw}$ are the final position in the *x* direction and swing time, respectively. $t_s$ is calculated by solving the following equation:

$$\frac{6x_f}{T_{sw}^5}t_s^5 - \frac{15x_f}{T_{sw}^4}t_s^4 + \frac{10x_f}{T_{sw}^3}t_s^3 - d_s = 0 \qquad (2)$$

The swing-leg trajectory in the *z* direction is divided into two trajectories, one for $t \leq t_s$ and the other for $t \geq t_s$ as follows

$$z = \begin{cases} \frac{6(h_s+\Delta h)}{t_s^5}t^5 - \frac{15(h_s+\Delta h)}{t_s^4}t^4 + \frac{10(h_s+\Delta h)}{t_s^3}t^3, & t \leq t_s \\ \frac{6(z_f-(h_s+\Delta h))}{(T_{sw}-t_s)^5}(t-t_s)^5 - \frac{15(z_f-(h_s+\Delta h))}{(T_{sw}-t_s)^4}(t-t_s)^4 \\ + \frac{10(z_f-(h_s+\Delta h))}{(T_{sw}-t_s)^3}(t-t_s)^3 + (h_s+\Delta h), & t > t_s \end{cases} \qquad (3)$$

Where $z_f$ and $h_s$ are the final position in the z direction and the height of the stair, respectively. For climbing down stairs, the wave gait with the procedure described in this section can be used but is skipped here for brevity.

## B. Spinning gait

Spinning gait is used when the robot is to rotate about the axis passing through the center of mass and perpendicular to the horizontal plane. In this gait, the robot body does not have any transitional movement, and only rotates about the *+z* or *−z* direction. Similar to the previous gait, the robot's feet at the beginning of the spinning gait should be positioned at appropriate positions with respect to the body reference frame (desired configuration) so that the robot can use its maximum workspace; however, each foot initially is often in the center of its workspace (initial configuration). Therefore, transition from the initial to desired configuration is required for accomplishing an optimal spinning gait

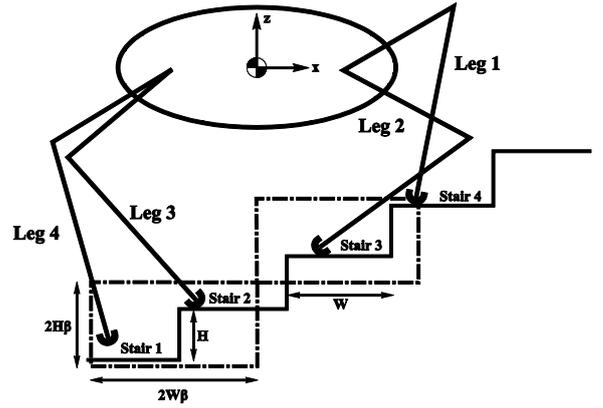

Fig. 4. The quadruped robot climbing up stairs.

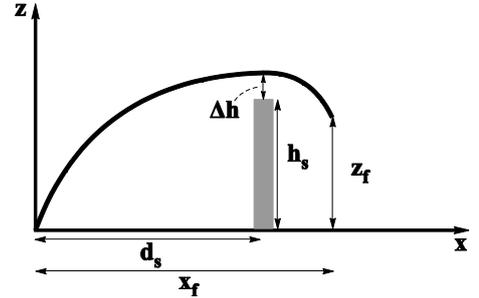

Fig. 5. The trajectory of the swing leg for stair climbing

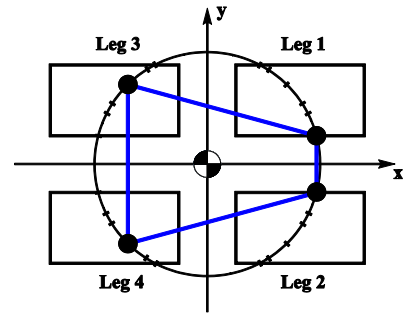

Fig. 6. The desired foot positions with respect to the body at the beginning of spinning gait.

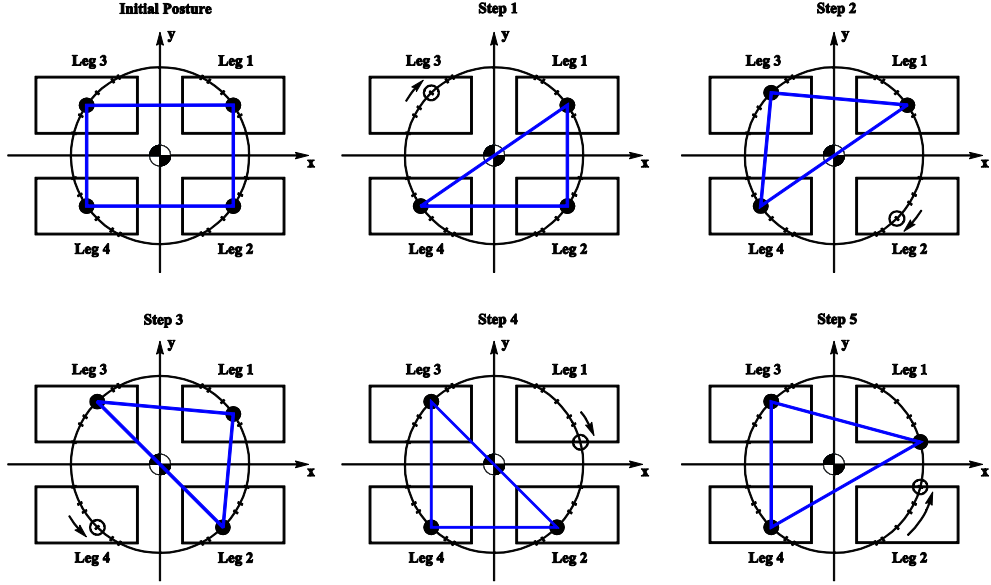

Fig. 7. Transition from the initial to desired configuration for the spinning gait.

For a counterclockwise spinning gait, the swing leg sequence is chosen as 1 – 3 – 4 – 2. The desired foothold positions with respect to the body at the beginning of the gait cycle are shown in Fig. 6. This configuration is chosen based on the swing leg sequence and the maximum use of each leg workspace. The transition from the initial to desired configuration is carried out in five steps as shown in Fig. 7. As seen, the final foothold positions at step $5$ are the same as those (desired configuration) shown in Fig. 6 and the projection of the center of mass lies inside or on the border of the support polygon at each step. This indicates a stable transition from the initial to desired configuration for the spinning gait.

In a spinning gait, each foot moves on a circular path. Here, we suppose that the foot circular path is chosen so that it passes through the center of each leg workspace. Depending on the dimensions of each leg workspace, the circular path can intersect the widths or lengths of the leg workspace. Here, we suppose that it intersects the lengths of the leg workspace. From Fig. 8, the parameter $\varphi$ is given by

$$\varphi = \frac{\pi}{2} - (\delta + \gamma) \tag{4}$$

Where

$$\delta = tan^{-1}\left(\frac{P_y - R_y}{\sqrt{P_x^2 + 2P_y R_y - R_y^2}}\right) \tag{5}$$

$$\gamma = tan^{-1}\left(\frac{\sqrt{P_x^2 - 2P_y R_y - R_y^2}}{P_y + R_y}\right) \tag{6}$$

All parameters in Eqs. (5) and (6) are shown in Fig. 8. The $x$ and $y$ components of the distance between points $A_1$ and $B_1$ are given by

$$s_x = \frac{\sqrt{P_x^2 + P_y^2}}{2}\{cos(\delta + \varphi) - cos(\delta)\}$$

$$s_y = \frac{\sqrt{P_x^2 + P_y^2}}{2}\{sin(\delta + \varphi) - sin(\delta)\} \tag{7}$$

For synthesizing the spinning gait, we divide the cycle time, T, into six intervals: $[t_0$-$t_1]$, $[t_1$-$t_2]$, $[t_2$-$t_3]$, $[t_3$-$t_4]$, $[t_4$-$t_5]$, $[t_5$-$t_6]$. Since the swing leg sequence for the counterclockwise spinning gait is chosen to be $1 – 3 – 4 – 2$, leg 1 is lifted from the ground at $t_0$, when its kinematic margin is zero, and is moved as $s_x$ and $s_y$ relative to the body in the x and y directions, and is touched the ground at $t_1$ (see Fig. 9). The angle $\Delta\Omega$ the body rotates with constant angular velocity during $[t_0$-$t_1]$ is

$$\Delta\Omega = \left(\frac{1}{\beta} - 1\right)\varphi \tag{8}$$

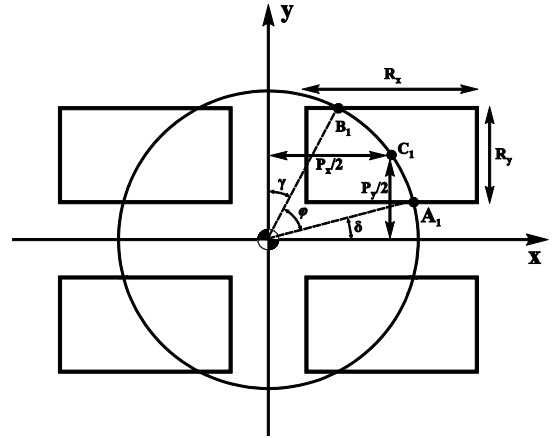

Fig. 8. Geometric configuration used for the spinning gait.

In the second interval, *[t₁-t₂]*, all legs are fixed on the ground while the body rotates with constant angular velocity as

$$\Delta\Omega' = \left(2 - \frac{3}{2\beta}\right)\varphi \qquad (9)$$

When the body rotates in the first and second intervals, leg *3* reaches its kinematic margin with respect to the body at $t_2$. Now leg *3* is lifted at $t_2$ and is moved as $s_x$ and $s_y$ relative to the body in *x* and *y* directions and is touched the ground at $t_3$. In this interval, the body also rotates as *ΔΩ* expressed by Eq. (8). In the next interval, *[t₃-t₄]*, leg *4* is lifted at $t_3$, is moved as $s_x$ and $s_y$ relative to the body in *x* and *y* directions, and is touched the ground at $t_4$. In this interval, the body also rotates as *ΔΩ*. In *[t₄-t₅]*, all legs are fixed on the ground and only the body rotates as *ΔΩ'* expressed by Eq. (9). In the last interval, *[t₅-t₆]*, leg *2* is lifted at $t_5$, is moved as $s_x$ and $s_y$ relative to the body in *x* and *y* directions, and is touched the ground at $t_6$. In this interval, the body rotates as *ΔΩ*. The relative leg motion for one period of the counterclockwise spinning gait is illustrated in Fig. 9. For a clockwise spinning gait, the swing-leg sequence is chosen as 1 – 2 – 4 – 3.

In order to make a stable transition from the wave gait to the spinning gait and vice versa, one simple way is to place each leg in the center of its workspace with respect to the body at the beginning of the new gait. Next, transition from the initial to desired configuration can be performed using the procedure described earlier. We use this approach in our simulations.

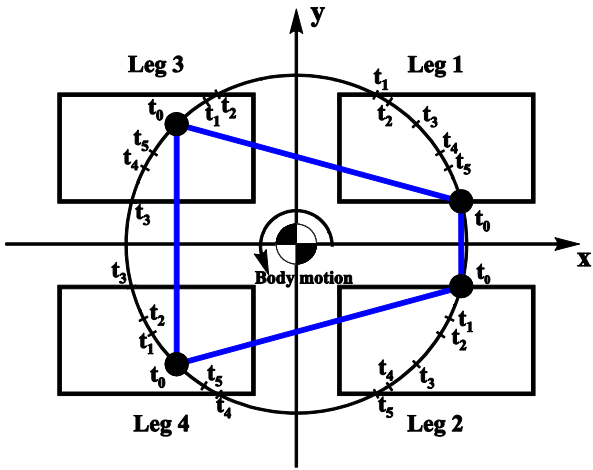

Fig. 9. Relative foot positioning in a counterclockwise spinning gait.

### III. RESULTS

For simulation, we consider a case study in which the robot first moves forward a few steps on the level ground, next ascends the stairs, then rotates *90* degrees, and finally descends the stairs. The robot model used here is the same as that used in our previous work [9, 10]. Fig. 10 illustrates the model traversing the stairs. Gait parameters used in the simulation are given in table I.

The trajectories of the body, leg *1* and leg *2* with respect to the ground are shown in Fig 11. The stability margin ($S_{SM}$) of the robot corresponding to the trajectories shown in Fig. 11 is illustrated in Fig. 12. As seen from this figure, the stability margin is greater than zero most of the time. In a few occasions, the stability margin is zero which is related to the transition from the initial to desired configuration. Therefore, the robot is able to stably perform the task without falling.

Table I. Gait parameters.

| $R_x(m)$ | $R_y$ (m) | $P_x$(m) | $P_y$ (m) | $\beta$ |
|---|---|---|---|---|
| 0.76 | 0.5 | 0.8 | 0.54 | 0.75 |
| $W(m)$ | $H(m)$ | $T$(sec) | $t_0$(sec) | $\Delta h(m)$ |
| 0.50 | 0.13 | 8 | 0 | 0.02 |

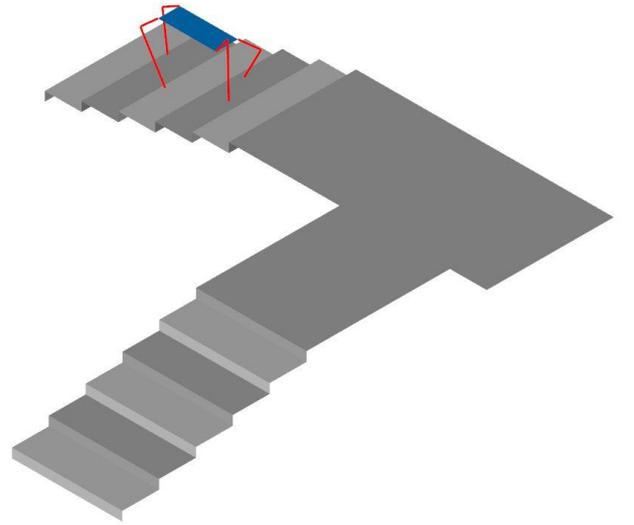

Fig. 10. The model of the quadruped robot during simulation.

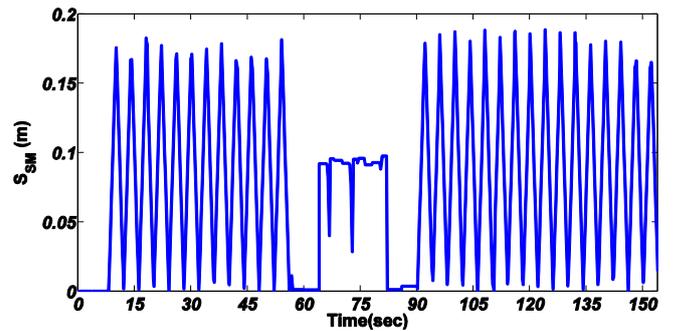

Fig. 12. Static stability margin versus time

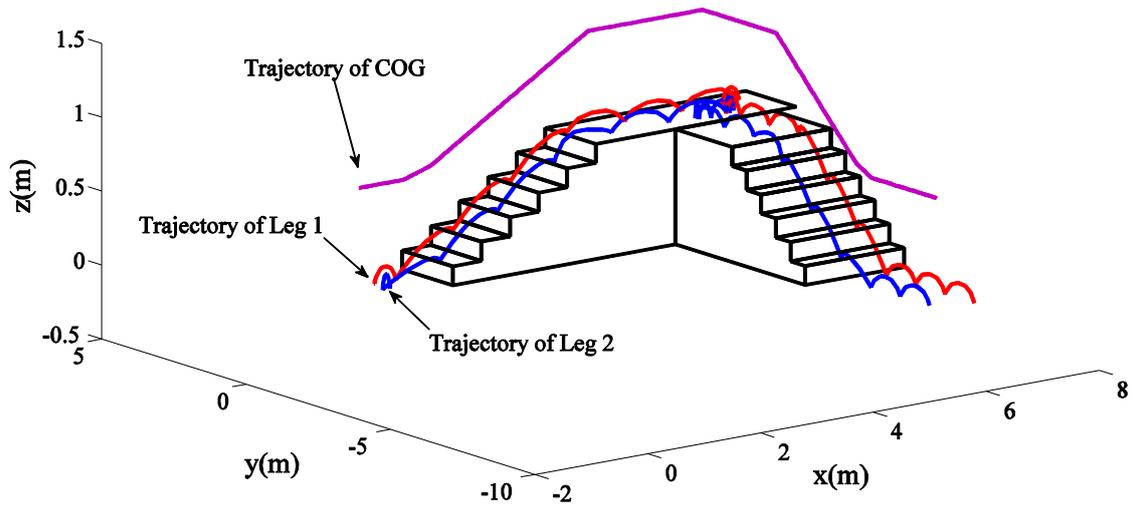

Fig. 11. Trajectories of the body, leg 1, and leg 2 with respect to the ground.

## VI. CONCLUSIONS

Here we demonstrated a gait planning algorithm that enabled the quadruped robot to climb up the stairs, rotate about its yaw axis, and climb down the stairs. To do so, a stable transition from the initial to desired configuration was performed for the wave gait based on the maximum use of leg workspace and minimum steps. Next, a wave gait formulation for climbing stairs was presented. A smooth leg swing trajectory was designed so that collision between the swing leg and stairs was avoided and the effect of impact during the foot strike with the stairs was reduced. Afterwards, similar to the wave gait, a stable transition from the initial to desired configuration was carried out for the spinning gait followed by the spinning gait formulation. The performance of the proposed gait planning algorithm was evaluated in simulation. It was shown that the robot could carry out the task while its stability was guaranteed. Our work has some limitations as follows. We did not use feedback control in our work, thus the robot is not robust to perturbations. We also ignored the mass of legs and supposed that the robot speed is low, therefore the effect of gravity dominates all other dynamic terms. Addressing these issues will be our future work.